\documentclass{article}
\usepackage{spconf,amsmath,graphicx}
\usepackage{cite}
\usepackage{siunitx}
\usepackage{amsmath,amssymb,amsfonts}
\usepackage{textcomp}
\usepackage{booktabs, makecell, multirow, tabularx}
\usepackage{cite}
\usepackage{url}
\usepackage{amsmath,amssymb,amsfonts}
\usepackage{algorithm}
\usepackage{graphicx}
\usepackage{changepage}
\usepackage{algpseudocode}
\usepackage{gensymb}
\usepackage[hidelinks]{hyperref} 
\usepackage{lipsum}

\usepackage{multicol}
\usepackage{subcaption}

\title{FARE: A Deep Learning-Based Framework for Radar-based Face Recognition and Out-of-distribution Detection}
\name{Sabri Mustafa Kahya$^{\star}$  \qquad Boran Hamdi Sivrikaya$^{\star}$ \qquad Muhammet Sami Yavuz$^{\star}$ \qquad Eckehard Steinbach$^{\star}$}

\address{  $^{\star}$Technical University of Munich,
School of Computation, Information and Technology,\\  Department of Computer Engineering, Chair of Media Technology,\\ Munich Institute of Robotics and Machine Intelligence (MIRMI)}

\begin{document}

\maketitle
\begin{abstract}
In this work, we propose a novel pipeline for face recognition and out-of-distribution (OOD) detection using short-range FMCW radar. The proposed system utilizes Range-Doppler and micro Range-Doppler Images. The architecture features a primary path (PP) responsible for the classification of in-distribution (ID) faces, complemented by intermediate paths (IPs) dedicated to OOD detection. The network is trained in two stages: first, the PP is trained using triplet loss to optimize ID face classification. In the second stage, the PP is frozen, and the IPs—comprising simple linear autoencoder networks—are trained specifically for OOD detection. Using our dataset generated with a 60 GHz FMCW radar, our method achieves an ID classification accuracy of 99.30\% and an OOD detection AUROC of 96.91\%.
\end{abstract}


\section{Introduction}
\label{sec:intro}

Short-range radars have gained significant popularity in various applications due to their cost-effectiveness, privacy-preserving nature, and robustness to environmental conditions. Unlike optical sensors, they do not capture identifiable visual data, making them ideal for privacy-sensitive environments. These features have led to their widespread use in applications such as human presence detection, activity classification, people counting, gesture recognition, and heartbeat estimation \cite{human_presence7,human_activity3,people_counting,gesture_recog,heartbeat_est}. In this study, we propose a face recognition and out-of-distribution (OOD) detection framework.

OOD detection  \cite{b9,b5,b27,b25} is essential for the secure and reliable deployment of deep learning models, as it prevents overconfident predictions on samples that lie away from the training data. Traditional classifiers can be easily misled by data outside their known classes, but OOD detection helps identify and reject unknown data. 

FARE functions both as a human face classifier for in-distribution (ID) faces and as an OOD detector for OOD faces. Its architecture simultaneously leverages Range-Doppler Images (RDIs) and micro Range-Doppler Images (micro-RDIs). As depicted in Figure 1, the network functions as a classifier along the primary path (PP). For OOD detection, the network utilizes intermediate paths (IPs) that branch off from each layer of the PP. Each IP plays a crucial role in filtering out OOD faces, which is the primary focus of our study. A typical OOD detector employs a scoring function along with a predefined threshold to classify test samples. If the score of a test sample falls below the threshold, it is classified as ID; if it exceeds the threshold, it is labeled as OOD. In our study, we derive the scores from the reconstruction errors produced by the IPs. Our key contributions: 
\begin{itemize}
\item  We introduce a unique architecture that incorporates both PP and IPs. The PP component is responsible for accurately classifying IDs, while the IPs are designed for OOD detection. Each IP consists of a simple linear autoencoder (encoder-decoder) network. Our training process is divided into two stages. In the first stage, we train the classifier, focusing on optimizing the PP for accurate face recognition. In the second stage, we freeze the PP and train only the reconstruction-based intermediate paths (IPs) for OOD detection. This approach allows us to achieve both a highly accurate human face classifier and an effective OOD detector.

\item We propose a novel loss function optimized in two stages: The first stage minimizes the classifier loss using a triplet loss in the PP for accurate ID classification. The second stage minimizes the reconstruction losses from the IPs for OOD detection, with six reconstruction losses corresponding to the IPs at the end of each layer. FARE achieves an average ID human face classification accuracy of 99.30\% and an average AUROC of 96.91\% for OOD detection. Also, FARE outperforms state-of-the-art (SOTA) OOD detection methods across commonly used evaluation metrics.

\end{itemize}

\section{Related Work}
\label{sec:related}

Several studies have explored radar-based face authentication. \cite{dnn-based} utilized a deep neural network (DNN) in combination with a $\SI{61}{\giga\hertz}$ mmWave radar to classify facial data from eight subjects. It achieved a classification accuracy of 92\%. In \cite{cnn-based}, researchers used CNNs with the same $\SI{61}{\giga\hertz}$ radar to identify faces from 30 cm away. They tested the model with and without cotton masks. The model was first trained on unmasked faces, then adjusted to recognize masked faces, and showed results for both situations. Further advancements include the deployment of a radar system with 32 transmit (Tx) and 32 receive (Rx) antennas, using a dense autoencoder for one-class verification across 200 individual faces. This method was enhanced through the integration of a convolutional autoencoder paired with a random forest classifier, improving upon the original system's performance \cite{32by32, imp32by32}. \cite{one-shot} introduced a one-shot learning approach based on a Siamese network architecture with a $\SI{61}{\giga\hertz}$ FMCW radar, which involved eight participants and reported a classification accuracy of 97.6\%.  While these studies have made significant improvements in classification accuracy, they overlook detecting OOD samples, which is essential for reliability and security.

The OOD detection concept was first introduced in \cite{b1}, where maximum softmax probabilities were used to distinguish OODs from IDs, based on the observation that OOD instances generally have lower softmax scores. To improve upon this, ODIN \cite{b2} utilized input perturbations and temperature scaling on logits to amplify the softmax scores of ID samples. MAHA \cite{b4} employed Mahalanobis distance for OOD detection, utilizing representations from intermediate network layers, a strategy also used by FSSD \cite{b6}. \cite{b7} proposed an energy-based method leveraging the $logsumexp$ function. ReAct \cite{b28} introduced a truncation method for activations in the penultimate layer, which is compatible with various existing OOD detection techniques. GradNorm \cite{b14} proposed a method to differentiate between ID and OOD samples by analyzing the norm of gradients. Additionally, MaxLogit and KL-matching methods from \cite{hendrycks2022scaling} offered alternatives, using maximum logit scores and minimum KL divergence, respectively. Approaches like OE \cite{b8} and OECC \cite{b10} used limited OOD samples during training, claiming that minimal OOD exposure helps distinguish between ID and OOD data.

Radar-based OOD detection has seen notable advancements. \cite{RB-OOD} used a $\SI{60}{\giga\hertz}$ FMCW radar to identify moving objects as OOD compared to walking people. MCROOD \cite{MCROOD} improved on this by incorporating a multi-class framework that also distinguishes between sitting and standing individuals. \cite{kahya2023hood} introduced a multi-encoder, multi-decoder network for concurrent human presence and OOD detection. \cite{kahya2023harood} focused on human activity classification while detecting OODs.

\vspace{-0.3cm}

\section{Radar System Design}
\vspace{-0.3cm}

Our radar chipset has one Tx and three Rx antennas. The Tx antenna emits chirp signals captured by the Rx antennas. These signals are mixed and low-pass filtered to produce the intermediate frequency (IF) signal, which is then digitized into raw Analog-to-Digital Converter (ADC) data, preparing it for subsequent processing.
Our model leverages both RDIs and micro-RDIs as inputs. For the RDI, \textbf{Range-FFT}  is applied to the fast-time signal to extract range, complemented by mean removal and the Moving Target Identification (\textbf{MTI}) process, which eliminates static targets. \textbf{Doppler-FFT} is then applied to the slow-time signal to capture phase variations. For the micro-RDI, \textbf{Range-FFT} is employed, stacking eight range spectrograms and applying mean removal to both fast and slow-time signals to minimize noise. \textbf{Sinc} filtering is used to enhance target detection, and \textbf{Doppler-FFT} along the slow-time dimension produces the micro-RDI. Lastly, we apply \textbf{E-RESPD} \cite{kahya2023hood} to improve the detection of facial movements in both RDIs.
\begin{figure*}[htbp]
\centerline{\includegraphics[width=0.69\linewidth]{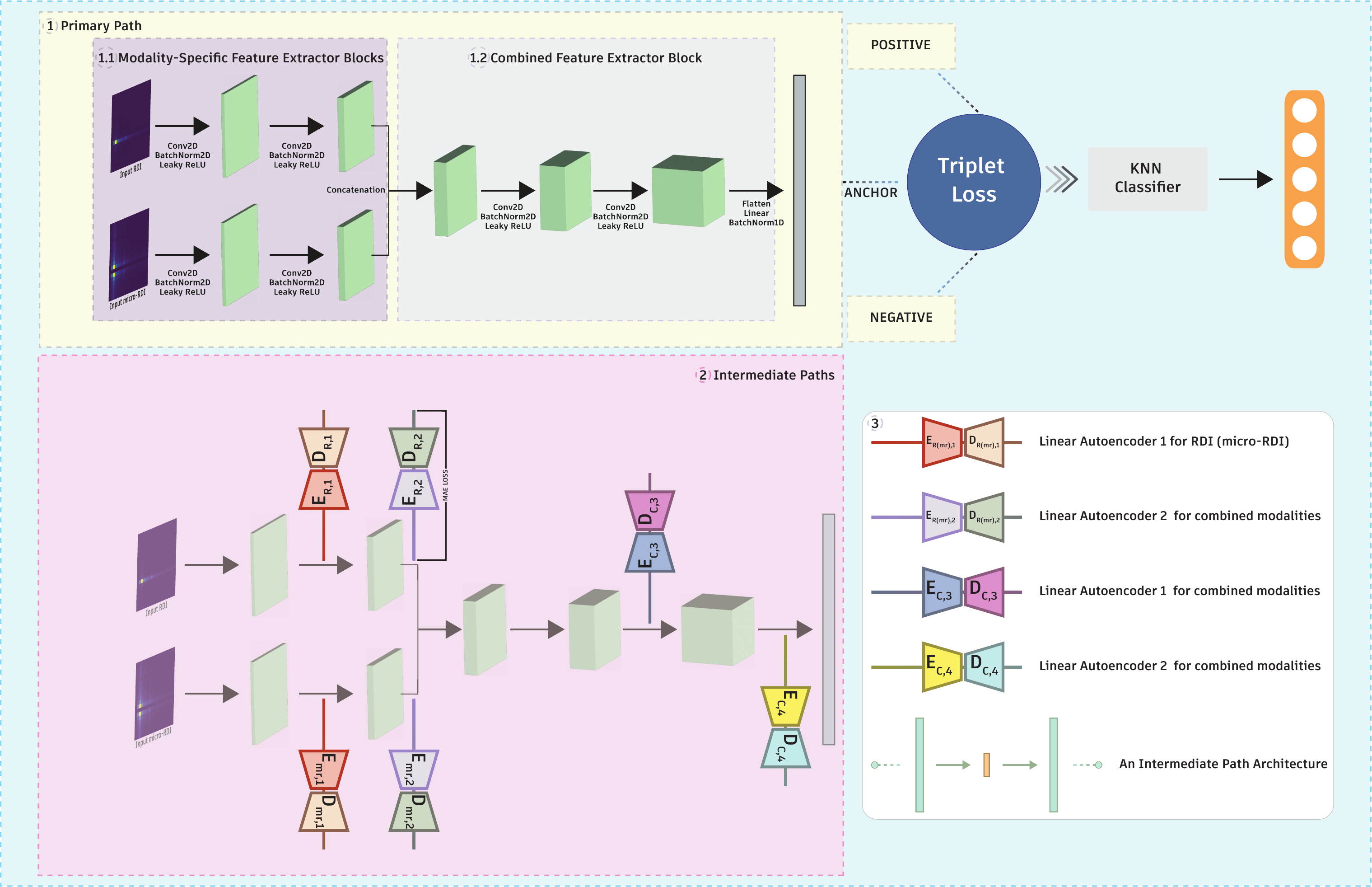}}
\caption{ \footnotesize This figure summarizes our architecture. The upper yellow section (1) represents the PP with initial modality-specific feature extractor blocks (1.1) and a combined feature extractor block (1.2). Minimized versions of the PP, also with a yellow background, represent the positive and negative samples, alongside the anchor for triplet training. A KNN classifier is used after extracting embeddings from the PP. The lower pink section (2) shows the intermediate paths (IPs). The PP remains frozen, as indicated in the figure, with training focused solely on the IPs. The IPs, which are linear encoder-decoder architectures, are located at the end of each layer of the PP and are responsible for OOD detection. The section shown in white and labeled 3 represents the navigator.}
\label{fig:pipline}
\end{figure*}
\vspace{-0.6cm}
\section{Problem Statement and FARE}
FARE simultaneously addresses face recognition and OOD detection. While many studies have explored face recognition with various sensors, most focus primarily on classification. Our study emphasizes eliminating OOD faces without compromising the accuracy of ID face classification.

The PP comprises initial modality-specific feature extractor blocks and a combined feature extractor block. The initial extractors, consisting of convolutional layers, are separately designed for RDIs and micro-RDIs, ensuring comprehensive feature extraction from both modalities. After these initial blocks, the intermediate features from both modalities are merged and passed to the combined feature extractor block. This block, containing convolutional and fully connected layers, generates final embeddings for triplet training, which enables the accurate classification of five ID faces. The PP is trained with triplet loss:
\begin{equation*}
\vspace{-0.2cm}
\begin{aligned}
\mathcal{L}_{PP}&  = \frac{1}{b} \sum_{i=1}^{b} \max(\|\textbf{t}^{(i)}_{e,a} -  \textbf{t}^{(i)}_{e,p}\|_2 - \| \textbf{t}^{(i)}_{e,a} - \textbf{t}^{(i)}_{e,n}\|_2 + m, 0),
\end{aligned}
\label{eq:loss_triplet}
\
\end{equation*}
where $\textbf{{t}}^{(i)}_{e,a},\textbf{{t}}^{(i)}_{e,p}$, and $\textbf{{t}}^{(i)}_{e,n}$ represent the triplet embeddings corresponding to the anchor (a), positive (p), and negative (n) samples, respectively. Here, $b$ denotes the batch size, and $m$ is the margin, which is set to 2.

Each IP uses a simple linear autoencoder architecture and is integrated at the end of each PP layer, including those within the initial modality-specific extractors. IPs are designed explicitly for OOD detection, intercepting OOD samples before they reach final classification. By positioning the IPs at the end of each PP layer, we leverage the unique information from each intermediate feature to effectively stop OOD samples. The IPs are trained with mean absolute error (MAE) loss. Our network comprises four main layers, but since the first two layers process two modalities separately, Layer 1 and Layer 2 each contain two sublayers. This results in a total of six layers, and consequently, we have six MAE losses:
\vspace{-0.1cm}
\begin{equation*}
\begin{aligned}
\mathcal{L}_{IP}  = \frac{1}{b} \sum_{j \in \{R,mR\}} \sum_{k=1}^{2} \sum_{i=1}^{b} (\textbf{I}^{(i)}_{j,k} - D_{j,k}(E_{j,k}(\textbf{I}^{(i)}_{j,k}))) + \\
\frac{1}{b} \sum_{z=3}^{4} \sum_{i=1}^{b} (\textbf{F}^{(i)}_{z} - D_{C,z}(E_{C,z}(\textbf{F}^{(i)}_{z}))),
\end{aligned}
\label{eq:loss_recons}
\
\end{equation*}

where $R$ and $mR$ represent RDI and micro-RDI modalities, respectively. $\textbf{I}^{(i)}_{j,k}$  represents an intermediate feature of an input with modality $j$ in layer $k$. $E_{j,k}$ is the encoder in the initial feature extractor block for modality $j$ in layer $k$, and similarly, $D_{j,k}$ represents the decoder in the same block. In the second part of the equation, $\textbf{F}^{(i)}_{z}$ is the intermediate feature of the combined modalities in layer $z$. $E_{C,z}$ and $D_{C,z}$ are the encoder and decoder, respectively, in layer $z$ of the combined feature extractor block, and $b$ denotes the batch size.

Training occurs in two stages using only ID face samples. First, the PP is trained with triplet loss using the Adamax \cite{kingma2014adam} optimizer. In the second stage, the PP is frozen, and the IPs are trained separately with MAE loss, also using Adamax. A simple K-nearest neighbors (KNN) algorithm is trained with the embeddings coming from PP for the final classification of the ID faces. Overall strategy yields a robust and accurate human face classifier and OOD detector.

\begin{table*}[htp]
\footnotesize
\centering
\caption{\footnotesize Comparison of our method with SOTA works across five ID classes (PER1/PER2/PER3/PER4/PER5) in terms of common OOD detection metrics. All values are shown in percentages. $\uparrow$ indicates that higher values are better, while $\downarrow$ indicates that lower values are better.}
\begin{tabular}{@{\extracolsep{\fill}}ccccc}
\hline
\textbf{Method} & \textbf{AUROC} & \textbf{AUPR\textsubscript{IN}} & \textbf{AUPR\textsubscript{OUT}} & \textbf{FPR95} \\ 
&  $\uparrow$
&  $\uparrow$
&  $\uparrow$
&  $\downarrow$ \\ \midrule

MSP\cite{b1} & 31.6/45.2/26.6/26.1/39.5 & 13.1/18.3/12.4/12.5/15.3 & 74.4/79.2/71.2/69.6/79.4 & 96.1/92.8/99.0/99.3/90.4 \\ 
ODIN\cite{b2} & 54.1/49.5/59.8/53.9/48.5 & 20.3/19.6/25.7/21.2/18.7 & 83.4/79.9/85.5/83.1/79.9 & 92.3/94.8/90.8/91.9/95.1 \\ 
ENERGY\cite{b7} & 48.1/31.7/29.4/67.2/56.2 & 17.1/13.9/13.1/29.9/19.9 & 81.3/70.6/69.7/90.3/86.6 & 94.3/98.3/99.6/72.6/79.4 \\ 
MAHA\cite{b4} & 46.2/42.6/61.5/64.4/49.2 & 16.5/16.6/24.9/27.2/18.1 & 80.4/76.4/87.1/88.1/81.1 & 94.1/96.8/86.5/84.5/93.0 \\ 
FSSD\cite{b6} & 55.4/42.5/51.0/24.6/31.9 & 19.5/19.5/17.9/12.2/13.6 & 85.6/77.9/83.8/69.7/75.7 & 85.4/92.9/88.5/98.3/91.9 \\ 
OE\cite{b8} & 79.2/29.7/83.3/67.6/58.8 & 35.0/13.6/45.6/26.0/21.0 & 94.8/73.8/95.6/90.9/88.2 & 54.1/91.0/63.4/65.8/68.6 \\ 
REACT\cite{b28} & 25.1/46.0/47.8/56.0/31.5 & 12.1/21.9/16.6/26.8/13.5 & 70.7/77.5/81.2/82.5/72.7 & 98.1/97.2/94.5/95.2/97.0 \\ 
GRADNORM\cite{b14} & 4.2/15.9/0.0/7.9/0.0 & 10.4/11.9/10.2/10.9/10.6 & 61.8/63.3/61.1/63.3/60.1 & 100/63.3/100/98.3/100 \\ 
MAXLOGIT\cite{hendrycks2022scaling} & 53.8/52.9/56.6/52.7/51.1 & 23.3/27.1/35.2/23.7/19.3 & 81.7/80.6/82.4/81.0/80.5 & 94.9/94.9/95.2/95.7/95.0 \\ 
KL\cite{hendrycks2022scaling} & 75.3/67.9/77.4/91.2/77.8 & 45.9/29.2/49.3/73.7/34.9 & 92.4/88.1/93.5/97.6/94.2 & 69.9/78.7/64.9/44.1/51.4 \\ 
\midrule
\midrule
\addlinespace
\textbf{FARE}&  \textbf{98.0/90.9/96.6/96.7/96.6} & \textbf{97.1/76.2/89.3/99.2/98.9} & \textbf{99.3/96.9/99.2/99.9/99.9} & \textbf{5.2/29.8/15.2/0.0/0.1} \\ \bottomrule

\end{tabular}
\label{tab:comparison}
\end{table*}

\subsection{OOD Detection \& Human Face Classification}

As is typical in OOD detection, we only use ID samples during training. The IPs, which are responsible for OOD detection, each consist of a simple linear autoencoder with a reconstruction-based architecture. Since the IPs are trained only on ID samples, we expect lower reconstruction errors for ID samples than OODs. We set a threshold that ensures 95\% of ID data are correctly classified. During testing, the sample is classified as OOD if the total reconstruction error or score from the IPs exceeds this predefined threshold. If the IPs do not filter the sample (its score is below the threshold, indicating it is ID), the sample continues through the PP and KNN for the classification of the ID face.

\section{Experiments and Results}
For this study, we created a face dataset using Infineon's BGT60TR13C $\SI{60}{\giga\hertz}$ FMCW radar sensor. Recordings were captured at a fixed distance of 25 cm from the sensor, with each session lasting 2 minutes. To maintain uniformity, participants were recorded without any facial accessories. Data was gathered across different days and at various times and featured a range of room environments to introduce background variability. The dataset comprises ID samples from four males and one female and OOD samples from nine males and two females. This resulted in a total of 80964 ID frames and 22458 OOD frames. 53536 frames of the ID data used for training, while 27428 for testing. The dataset is accessible at here\footnote{\href{https://syncandshare.lrz.de/getlink/fiJ1ZPazuTkCGgUGnQAGVG/}{https://syncandshare.lrz.de/getlink/fiJ1ZPazuTkCGgUGnQAGVG/}}, with all participants providing written consent for their involvement.
 This work specifically addresses the OOD detection problem in facial recognition systems, which is why we use a large number of OOD faces. We focus on a smaller number of ID faces, making our system highly suitable for smart home environments. For instance, our pipeline can be integrated into a smart TV to provide personalized content for parents and children while filtering unknown faces, either displaying default content or keeping the TV off.

We employed standard evaluation metrics. For OOD detection, we utilized four key metrics. \textbf{AUROC} represents the area under the receiver operating characteristic (ROC) curve. \textbf{AUPR\textsubscript{IN/OUT}} measures the area under the precision-recall curve when ID/OOD samples are considered as positive cases. \textbf{FPR95} denotes the false positive rate (FPR) when the true positive rate (TPR) reaches 95\%. 

\begin{figure}[ht]

    \centering

    \includegraphics[width=0.65\columnwidth]{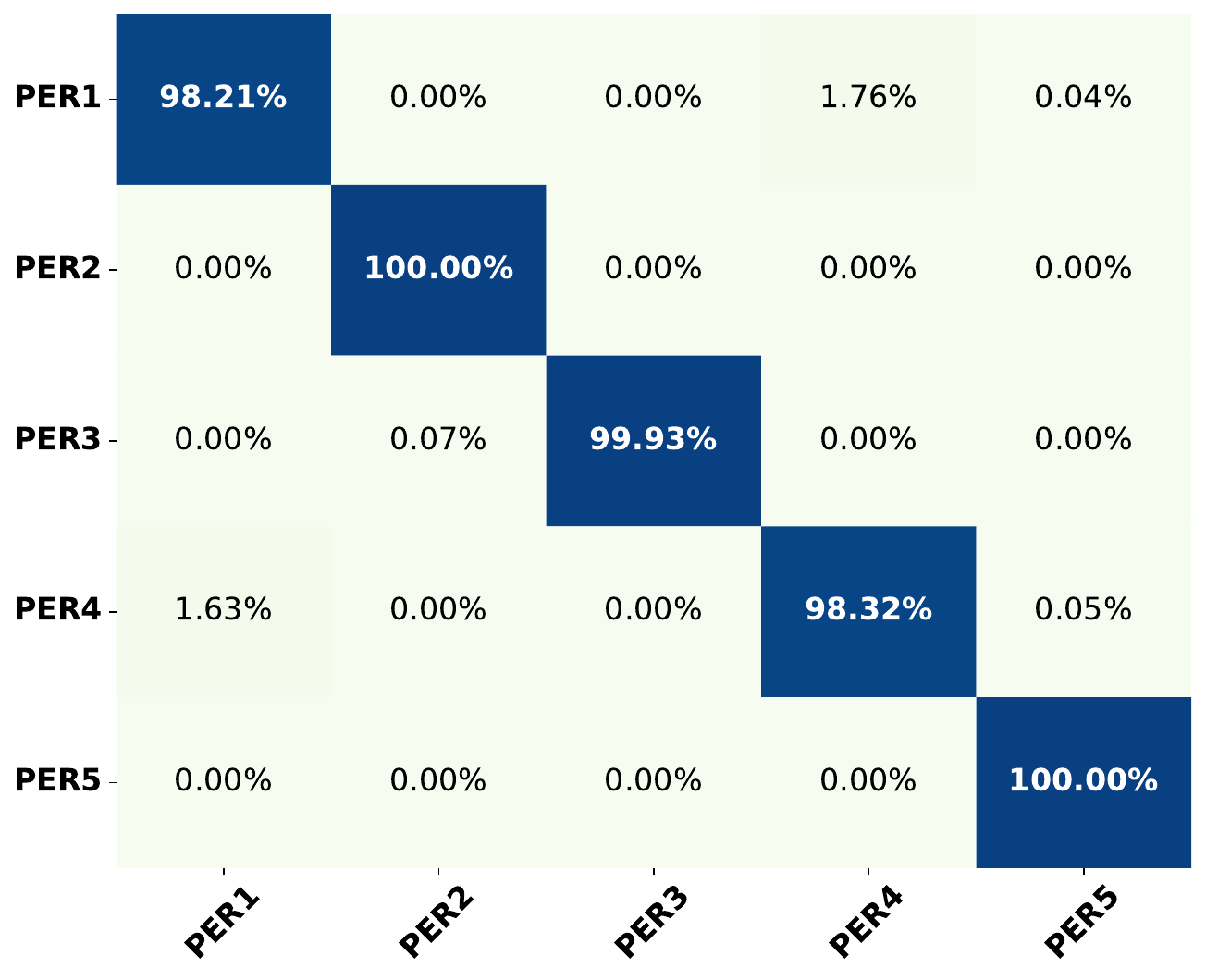}
    \label{fig:conf_matrix_FOOD}
     \caption{ \footnotesize Confusion matrix to show the classification performance of FARE.}
     \label{fig:conf_matrices}
\end{figure}

To evaluate the performance of FARE, we trained a ResNet34 \cite{resnet} backbone in a multi-class classification setup. It achieves slightly lower accuracy in human face classification at 99.15\% compared to FARE (\textbf{99.30\%}). Even though it achieves acceptable human face classification accuracy, it cannot address OOD detection. To assess the OOD detection capabilities of FARE relative to SOTA methods, we used the same pre-trained ResNet model to employ OOD detection works. We applied 10 different OOD detectors to the model to benchmark their performance against FARE, with a detailed comparison presented in Table \ref{tab:comparison}. The same training and testing datasets used for FARE were employed to ensure consistent comparison conditions with SOTA methods. Also, Figure \ref{fig:conf_matrices} displays the confusion matrix for FARE.

\vspace{-0.4cm}
\subsection{Ablation}
\vspace{-0.2cm}

We conducted an ablation study to evaluate the impact of incorporating Intermediate Paths (IPs) at different layers of the Primary Path (PP) on OOD detection. The study aims to demonstrate how utilizing intermediate feature representations with IPs enhances OOD detection performance.
Table \ref{tab:ablation_layers} highlights that employing IPs at more layers improves OOD detection performance, showing the positive effect of leveraging intermediate feature representations in our architecture. Our network has four main layers, but includes six IPs because the first two layers process two modalities separately. This means that both Layer 1 and Layer 2 each contain two sub-layers. For a visual reference, please see Figure \ref{fig:pipline}.

\begin{table}[ht]
\footnotesize
    \caption{ \footnotesize Ablation study showing the impact of using multiple IPs at the end of each PP layer for OOD detection. For example, ``Layers 1-2" indicates the use of IPs only at Layers 1 and 2 for OOD detection.}
    \centering
    \begin{tabular}{@ {\extracolsep{5pt}} ccccc}
    \toprule
    
    \centering
        IPs at & AUROC & AUPR\textsubscript{IN} & AUPR\textsubscript{OUT} & FPR95  \\
    &  $\uparrow$
    &  $\uparrow$
    &  $\uparrow$
    &  $\downarrow$\\
   
    \midrule
       Layer 1 & 89.83 &  91.34 & 88.63 & 25.0\\
         Layer 1-2  & 92.71 &  94.15 & 91.40 & 25.69\\ 
         Layer 1-2-3 & 94.84 & 96.03& 93.36& 22.97\\

        \textbf{Layer 1-2-3-4}  &  \textbf{96.91} & \textbf{97.74} & \textbf{95.43}& \textbf{19.53}\\ 
    
    \bottomrule
    \end{tabular}
    \label{tab:ablation_layers}
\end{table}
\vspace{-0.5cm}
\section{Conclusion}
\vspace{-0.2cm}

This work introduced FARE, a novel pipeline for face recognition and OOD detection using short-range FMCW radar. FARE combines ID face classification with accurate OOD detection by leveraging RDIs and micro-RDIs. The architecture, featuring PP for face recognition and IPs for OOD detection, demonstrated its effectiveness through two-stage training. Experimental results using our $\SI{60}{\giga\hertz}$ FMCW radar dataset showed high ID classification accuracy and OOD detection performance. This method advances OOD detection while offering a reliable solution for smart home applications.


\newpage

\footnotesize
\begin{thebibliography}{10}

\bibitem{human_presence7}
Prateek Nallabolu, Li~Zhang, Hong Hong, and Changzhi Li,
\newblock ``Human presence sensing and gesture recognition for smart home applications with moving and stationary clutter suppression using a 60-ghz digital beamforming fmcw radar,''
\newblock {\em IEEE Access}, vol. 9, pp. 72857--72866, 2021.

\bibitem{human_activity3}
Thomas Stadelmayer, Markus Stadelmayer, Avik Santra, Robert Weigel, and Fabian Lurz,
\newblock ``Human activity classification using mm-wave fmcw radar by improved representation learning,''
\newblock in {\em Proceedings of the 4th ACM Workshop on Millimeter-Wave Networks and Sensing Systems}, New York, NY, USA, 2020, mmNets'20, Association for Computing Machinery.

\bibitem{people_counting}
Cem~Yusuf Aydogdu, Souvik Hazra, Avik Santra, and Robert Weigel,
\newblock ``Multi-modal cross learning for improved people counting using short-range fmcw radar,''
\newblock in {\em 2020 IEEE International Radar Conference (RADAR)}, 2020, pp. 250--255.

\bibitem{gesture_recog}
Souvik Hazra and Avik Santra,
\newblock ``Robust gesture recognition using millimetric-wave radar system,''
\newblock {\em IEEE Sensors Letters}, vol. 2, no. 4, pp. 1--4, 2018.

\bibitem{heartbeat_est}
Muhammad Arsalan, Avik Santra, and Christoph Will,
\newblock ``Improved contactless heartbeat estimation in fmcw radar via kalman filter tracking,''
\newblock {\em IEEE Sensors Letters}, vol. 4, no. 5, pp. 1--4, 2020.

\bibitem{b9}
Qing Yu and Kiyoharu Aizawa,
\newblock ``Unsupervised out-of-distribution detection by maximum classifier discrepancy,''
\newblock in {\em IEEE/CVF International Conference on Computer Vision (ICCV)}, 2019.

\bibitem{b5}
Chandramouli~Shama Sastry and Sageev Oore,
\newblock ``Detecting out-of-distribution examples with {G}ram matrices,''
\newblock in {\em International Conference on Machine Learning (ICML)}, 2020.

\bibitem{b27}
Jingkang Yang, Haoqi Wang, Litong Feng, Xiaopeng Yan, Huabin Zheng, Wayne Zhang, and Ziwei Liu,
\newblock ``Semantically coherent out-of-distribution detection,''
\newblock in {\em IEEE International Conference on Computer Vision (ICCV)}, 2021.

\bibitem{b25}
Haoqi Wang, Zhizhong Li, Litong Feng, and Wayne Zhang,
\newblock ``Vim: Out-of-distribution with virtual-logit matching,''
\newblock {\em arXiv}, 2022.

\bibitem{dnn-based}
Hae-Seung Lim, Jaehoon Jung, Jae-Eun Lee, Hyung-Min Park, and Seongwook Lee,
\newblock ``Dnn-based human face classification using 61 ghz fmcw radar sensor,''
\newblock {\em IEEE Sensors Journal}, vol. 20, no. 20, pp. 12217--12224, 2020.

\bibitem{cnn-based}
J.~Kim, J.‐E Lee, H.‐S Lim, and S.~Lee,
\newblock ``Face identification using millimetre-wave radar sensor data,''
\newblock {\em Electronics Letters}, vol. 56, 08 2020.

\bibitem{32by32}
Eran Hof, Amichai Sanderovich, Mohammad Salama, and Evyatar Hemo,
\newblock ``Face verification using mmwave radar sensor,''
\newblock in {\em 2020 International Conference on Artificial Intelligence in Information and Communication (ICAIIC)}, 2020, pp. 320--324.

\bibitem{imp32by32}
Muralidhar~Reddy Challa, Abhinav Kumar, and Linga~Reddy Cenkeramaddi,
\newblock ``Face recognition using mmwave radar imaging,''
\newblock in {\em 2021 IEEE International Symposium on Smart Electronic Systems (iSES)}, 2021, pp. 319--322.

\bibitem{one-shot}
Ha-Anh Pho, Seongwook Lee, Vo-Nguyen Tuyet-Doan, and Yong-Hwa Kim,
\newblock ``Radar-based face recognition: One-shot learning approach,''
\newblock {\em IEEE Sensors Journal}, vol. 21, no. 5, pp. 6335--6341, 2021.

\bibitem{b1}
Dan Hendrycks and Kevin Gimpel,
\newblock ``A baseline for detecting misclassified and out-of-distribution examples in neural networks,''
\newblock in {\em International Conference on Learning Representations (ICLR)}, 2017.

\bibitem{b2}
Shiyu Liang, Yixuan Li, and R.~Srikant,
\newblock ``Enhancing the reliability of out-of-distribution image detection in neural networks,''
\newblock in {\em International Conference on Learning Representations (ICLR)}, 2018.

\bibitem{b4}
Kimin Lee, Kibok Lee, Honglak Lee, and Jinwoo Shin,
\newblock ``A simple unified framework for detecting out-of-distribution samples and adversarial attacks,''
\newblock in {\em International Conference on Neural Information Processing Systems (NeurIPS)}, 2018.

\bibitem{b6}
Haiwen Huang, Zhihan Li, Lulu Wang, Sishuo Chen, Bin Dong, and Xinyu Zhou,
\newblock ``Feature space singularity for out-of-distribution detection,''
\newblock in {\em Proceedings of the Workshop on Artificial Intelligence Safety 2021 (SafeAI 2021)}, 2021.

\bibitem{b7}
Weitang Liu, Xiaoyun Wang, John Owens, and Yixuan Li,
\newblock ``Energy-based out-of-distribution detection,''
\newblock in {\em Advances in Neural Information Processing Systems (NeurIPS)}, 2020.

\bibitem{b28}
Yiyou Sun, Chuan Guo, and Yixuan Li,
\newblock ``React: Out-of-distribution detection with rectified activations,''
\newblock in {\em Advances in Neural Information Processing Systems (NeurIPS)}, 2021.

\bibitem{b14}
Rui Huang, Andrew Geng, and Yixuan Li,
\newblock ``On the importance of gradients for detecting distributional shifts in the wild,''
\newblock in {\em Advances in Neural Information Processing Systems (NeurIPS)}, 2021.

\bibitem{hendrycks2022scaling}
Dan Hendrycks, Steven Basart, Mantas Mazeika, Andy Zou, Joe Kwon, Mohammadreza Mostajabi, Jacob Steinhardt, and Dawn Song,
\newblock ``Scaling out-of-distribution detection for real-world settings,''
\newblock {\em International Conference on Machine Learning (ICML)}, 2022.

\bibitem{b8}
Dan Hendrycks, Mantas Mazeika, and Thomas Dietterich,
\newblock ``Deep anomaly detection with outlier exposure,''
\newblock in {\em International Conference on Learning Representations (ICLR)}, 2019.

\bibitem{b10}
Aristotelis-Angelos Papadopoulos, Mohammad~Reza Rajati, Nazim Shaikh, and Jiamian Wang,
\newblock ``Outlier exposure with confidence control for out-of-distribution detection,''
\newblock {\em Neurocomputing}, vol. 441, pp. 138--150, 2021.

\bibitem{RB-OOD}
Sabri~Mustafa Kahya, Muhammet~Sami Yavuz, and Eckehard Steinbach,
\newblock ``Reconstruction-based out-of-distribution detection for short-range fmcw radar,''
\newblock in {\em 2023 31st European Signal Processing Conference (EUSIPCO)}, 2023, pp. 1350--1354.

\bibitem{MCROOD}
Sabri~Mustafa Kahya, Muhammet Sami~Yavuz, and Eckehard Steinbach,
\newblock ``Mcrood: Multi-class radar out-of-distribution detection,''
\newblock in {\em ICASSP 2023 - 2023 IEEE International Conference on Acoustics, Speech and Signal Processing (ICASSP)}, 2023, pp. 1--5.

\bibitem{kahya2023hood}
Sabri~Mustafa Kahya, Muhammet~Sami Yavuz, and Eckehard Steinbach,
\newblock ``Hood: Real-time robust human presence and out-of-distribution detection with low-cost fmcw radar,''
\newblock {\em arXiv}, 2023.

\bibitem{kahya2023harood}
Sabri~Mustafa Kahya, Muhammet Sami~Yavuz, and Eckehard Steinbach,
\newblock ``Harood: Human activity classification and out-of-distribution detection with short-range fmcw radar,''
\newblock in {\em ICASSP 2024 - 2024 IEEE International Conference on Acoustics, Speech and Signal Processing (ICASSP)}, 2024, pp. 6950--6954.

\bibitem{kingma2014adam}
Diederik~P. Kingma and Jimmy Ba,
\newblock ``Adam: A method for stochastic optimization,''
\newblock {\em arXiv preprint arXiv:1412.6980}, 2014.

\bibitem{resnet}
Kaiming He, Xiangyu Zhang, Shaoqing Ren, and Jian Sun,
\newblock ``Deep residual learning for image recognition,''
\newblock in {\em IEEE Conference on Computer Vision and Pattern Recognition (CVPR)}, 2016.

\end{thebibliography}
\end{document}